ORIGINAL RESEARCH

# Clustering with Minimum Spanning Trees: How Good Can It Be?

**Marek Gagolewski**[1,2] · **Anna Cena**[2] · **Maciej Bartoszuk**[3] · **Łukasz Brzozowski**[2]

## Abstract

Minimum spanning trees (MSTs) provide a convenient representation of datasets in numerous pattern recognition activities. Moreover, they are relatively fast to compute. In this paper, we quantify the extent to which they are meaningful in low-dimensional partitional data clustering tasks. By identifying the upper bounds for the agreement between the best (oracle) algorithm and the expert labels from a large battery of benchmark data, we discover that MST methods can be very competitive. Next, we review, study, extend, and generalise a few existing, state-of-the-art MST-based partitioning schemes. This leads to some new noteworthy approaches. Overall, the Genie and the information-theoretic methods often outperform the non-MST algorithms such as K-means, Gaussian mixtures, spectral clustering, Birch, density-based, and classical hierarchical agglomerative procedures. Nevertheless, we identify that there is still some room for improvement, and thus the development of novel algorithms is encouraged.

## 1 Introduction

Clustering methods aim at finding some meaningful partitions of a given dataset in a purely unsupervised manner. They have proven useful in abundant practical applications; e.g., in

✉ Marek Gagolewski
marek.gagolewski@pw.edu.pl

Anna Cena
anna.cena@pw.edu.pl

Maciej Bartoszuk
maciej.bartoszuk@qed.pl

Łukasz Brzozowski
lukasz.brzozowski3.dokt@pw.edu.pl

[1] Systems Research Institute, Polish Academy of Sciences, ul. Newelska 6, 01-447 Warsaw, Poland

[2] Warsaw University of Technology, Faculty of Mathematics and Information Science, ul. Koszykowa 75, 00-662 Warsaw, Poland

[3] QED Software, ul. Miedziana 3A, 00-814 Warsaw, Poland

medical, environmental, and earth sciences or signal processing (Guo et al., 2023; Hwang et al., 2023; Zhao et al., 2023; Zhou et al., 2023). Up to this date, many partitional clustering approaches have been proposed; for an overview, see the works by Wierzchoń and Kłopotek (2018), Blum et al. (2020), or Jaeger and Banks (2023). Methods to assess their usefulness include internal (Arbelaitz et al., 2013; Gagolewski et al., 2021; Halkidi et al., 2001; Jaskowiak et al., 2022; Maulik & Bandyopadhyay, 2002; Milligan & Cooper, 1985; Xu et al., 2020) and external cluster validity measures (Horta & Campello, 2015; Rezaei & Fränti, 2016; van der Hoef & Warrens, 2019; Wagner & Wagner, 2006; Warrens & van der Hoef, 2022) that were applied on various kinds of benchmark data (Dua & Graff, 2021; Fränti & Sieranoja, 2018; Gagolewski, 2022; Graves & Pedrycz, 2010; Thrun & Ultsch, 2020).

Given a dataset $\mathbf{X} = \{\mathbf{x}_1, \ldots, \mathbf{x}_n\}$ with $n$ points in $\mathbb{R}^d$, the space of all its possible $k$-partitions, $\mathcal{X}_k$, is very large. Namely, the number of divisions of $\mathbf{X}$ into $k \geq 2$ nonempty, mutually disjoint clusters is equal to the Stirling number of the second kind, $\left\{ {n \atop k} \right\} = O(k^n)$.

In practice, to make the identification of clusters more tractable, many algorithms tend to construct simpler representations (samples) of the search space. For instance, in the well-known K-means algorithm by Lloyd (1957), we iteratively seek $k$ (continuous) cluster centroids so that a point's cluster belongingness can be determined through the proximity thereto. In hierarchical agglomerative algorithms, we start with $n$ singletons, and then keep merging pairs of clusters that optimise some utility measure, e.g., the minimal or average intra-cluster distances, until we obtain $k$ point groups (Murtagh, 1983; Müllner, 2011; Szekely & Rizzo, 2005). In divisive schemes, on the other hand, we start with one cluster consisting of all the points, and keep splitting it into smaller and smaller chunks based on some criteria.

From the perspective of clustering tractability, different spanning trees offer a very attractive representation of data. Any spanning tree representing a dataset with $n$ points has $n - 1$ edges. If we remove $k - 1$ of them, we will obtain $k$ connected components which we can treat as clusters; compare Fig. 1. This reduces the size of the sample of the search space to explore greatly: down to $\binom{n-1}{k-1} = O(n^{k-1})$, and usually $k \ll n$. Some heuristics, such as greedy approaches, allow for further simplifications.

In particular, the minimum spanning tree (MST; the shortest dendrite) with respect to the Euclidean metric minimises the sum of pairwise distances. More formally, given an undirected weighted graph $G = (V, E, W)$ with $V = \{1, \ldots, n\}$, $E = \{\{u, v\}, u < v\}$, and $W(\{u, v\}) = \|\mathbf{x}_u - \mathbf{x}_v\|$, the minimum spanning tree $T = \mathrm{MST}(G) = (V, E', W')$, $E' \subset E$, $W' = W|_{E'}$ is a connected tree spanning $V$ with $E'$ minimising $\sum_{\{u,v\} \in E'} W(\{u, v\})$. Note that in this paper, we assume that an MST is always unique. This can be assured by adding, e.g., a tiny amount of noise to the points' coordinates in the case where there are tied distances.

MSTs are fast to compute. For general metrics, they require $O(n^2)$ time; see the classic algorithms by Borůvka (1926), Jarník (1930)[1], and Kruskal (1956). In small-dimensional Euclidean spaces, further speed-ups are achievable. For instance, March et al. (2010) prove that the time complexity is $\Omega(n \log n)$ for $d = 2$. An approximate MST can be computed as well (Naidan et al., 2019; Zhong et al., 2015).

Applications of MST-based clustering algorithms are plentiful: they are useful, e.g., in studying functional relationships of genes or recognising patterns in images (Xu et al., 2002; Yin & Liu, 2009). Unlike in the K-means algorithm which is connected to Voronoi's diagrams, in our setting, the clusters do not necessarily have to be convex. MST-based clustering methods allow for detecting well-separated, relatively dense clusters of arbitrary shapes, e.g., spirals, connected line segments, or blobs; see Fig. 2 for an example. This is true at least in low-dimensional spaces, albeit not necessarily in general ones due to the so-called curse of dimensionality (Blum et al., 2020). Also, note that some statistical properties of MSTs were studied by Di Gesu and Sacco (1983) and Jackson and Read (2010a, b).

[1] The Jarník method is more widely known as an algorithm by Prim (1957); for some historical notes, see the papers by Graham and Hell (1985), Zhong et al. (2015), and Gower and Ross (1969).

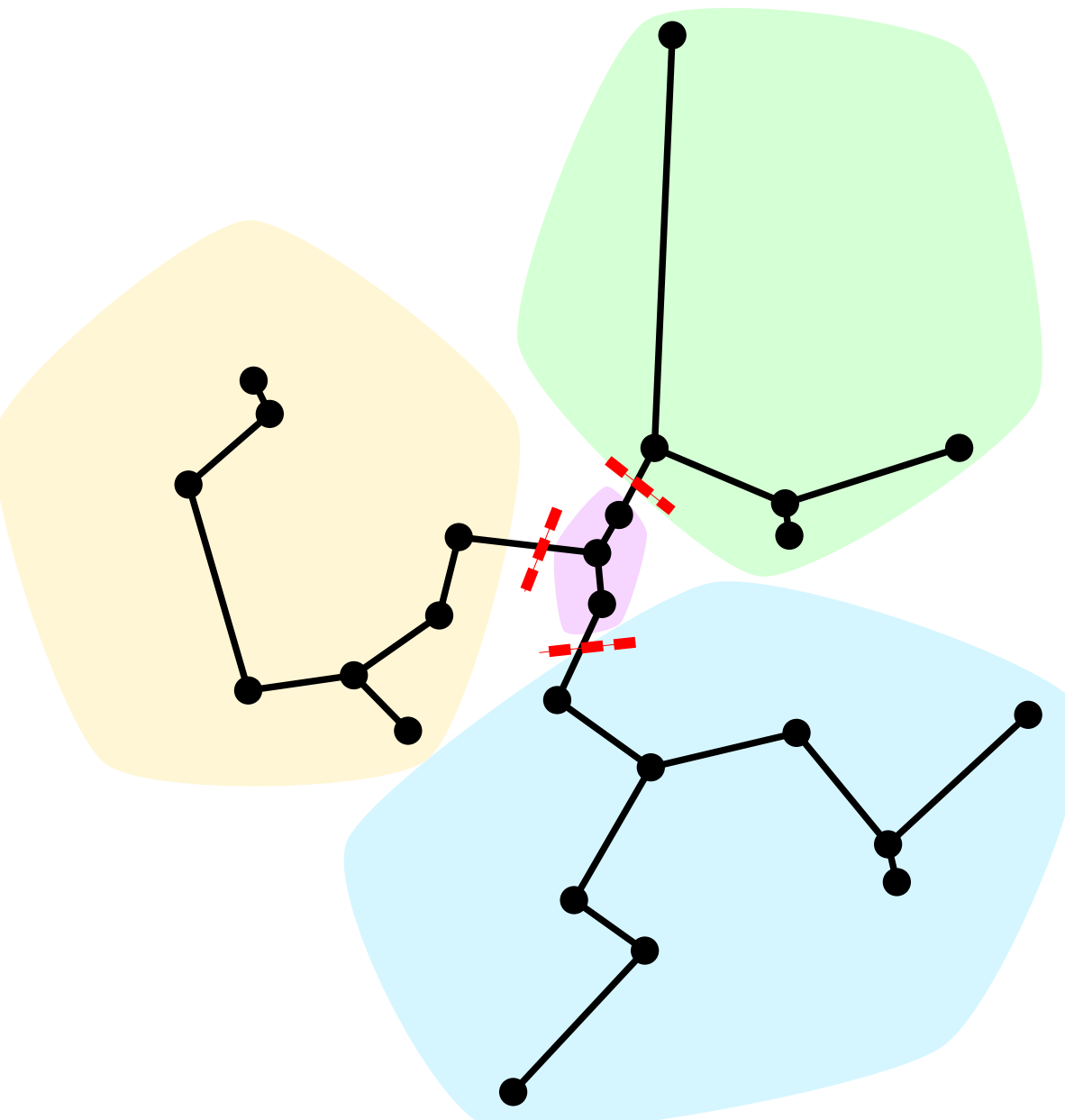

**Fig. 1** Removing three edges from a spanning tree gives four connected components, which we can treat as separate clusters

In this work, we aim to review, unify, and extend a large number of existing approaches to clustering based on MSTs that yield a specific number of clusters, $k$. Studying an extensive set of benchmark data of low-dimensionality (so that the results can be assessed visually) and moderate sizes ($n < 10{,}000$), we determine which of them performs best. Furthermore, we quantify how well particular MST-based methods perform in general, and determine whether they are competitive relative to other popular clustering procedures.

We have structured the current article as follows. Section 2 reviews existing MST-based methods and introduces a few of their generalisations, which include the divisive and agglomerative schemes optimising different cluster validity measures (with or without additional constraints). In Sect. 3, we answer the question of whether MSTs can provide us with a meaningful representation of the cluster structure in a large battery of synthetic benchmark datasets. Then, we pinpoint the best-performing algorithms, and compare them with state-of-the-art non-MST approaches. Section 4 concludes the paper and suggests some topics for further research.

## 2 Methods

Table 1 lists all the methods that we examine in this study. Let us first describe the MST-based approaches in detail.

SIPU/spiral WUT/smile WUT/x1
FCPS/wingnut SIPU/pathbased Other/chameleon_t8_8k
Graves/parabolic WUT/twosplashes WUT/olympic

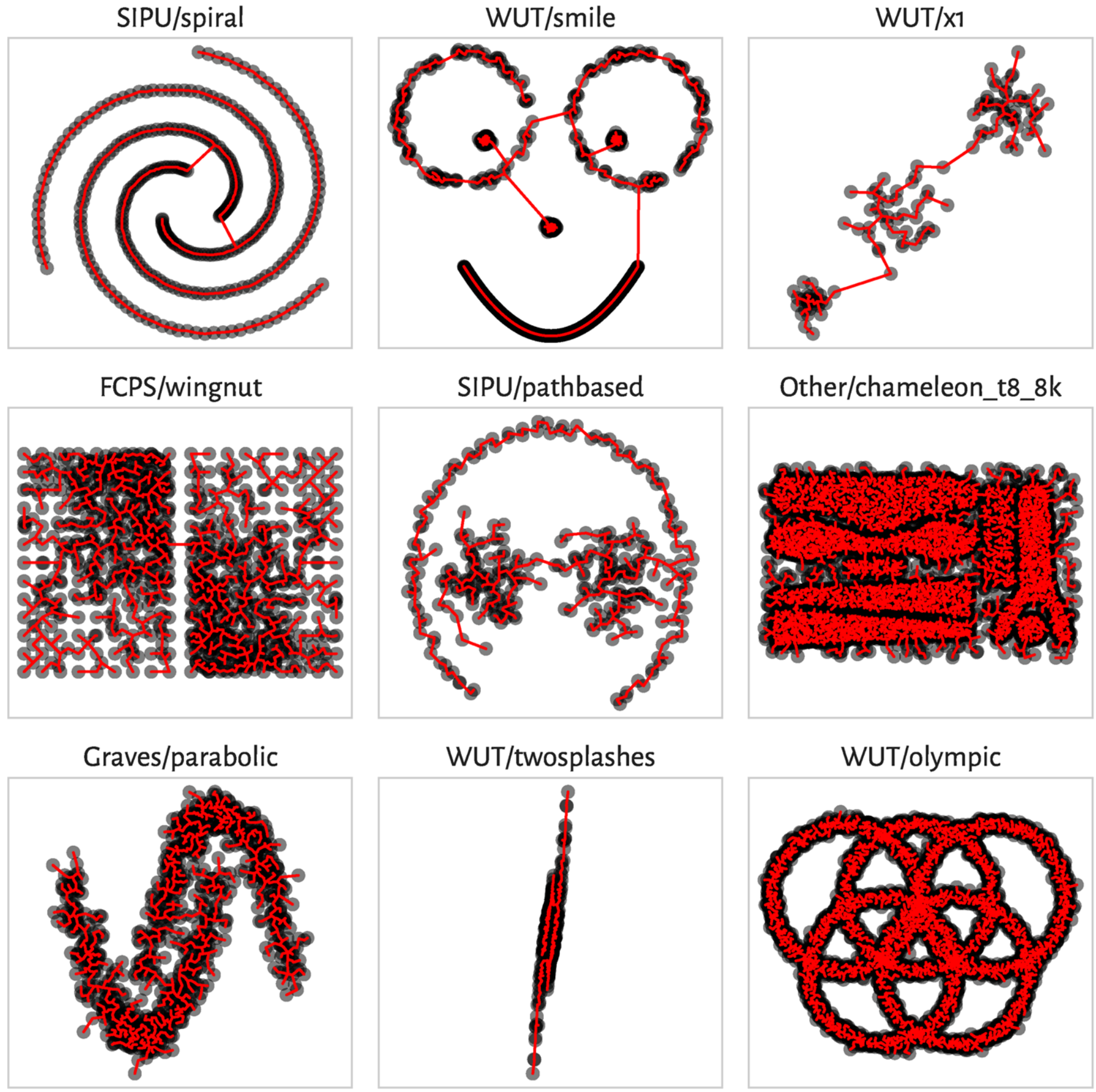

**Fig. 2** Example benchmark datasets (see Table 2) and their Euclidean minimum spanning trees. MSTs often lead to meaningful representations of well-separable clusters of arbitrary shapes, at least in low-dimensional spaces. Some datasets, however, might be difficult for all the algorithms

## 2.1 Divisive Algorithms over MSTs

Perhaps the most widely known MST-based method is the classic single linkage scheme (Wrocław Taxonomy, dendrite method, nearest neighbour clustering). It was proposed by Polish mathematicians Florek, Łukasiewicz, Perkal, Steinhaus, and Zubrzycki (1951). Gower and Ross (1969) noted that it can be computed using a divisive scheme: remove the $k-1$ edges of the maximal lengths from the MST, and consider the resulting connected components as separate clusters.

Another divisive algorithm over MSTs was studied by Caliński and Harabasz (1974). They optimised the total within-cluster sum of squares[2] (see Eq. 1 below). To avoid multiplying entities beyond necessity, let us introduce it in a more general language.

[2] It is the same objective function as in the K-means algorithm; the authors provided it as an alternative to the agglomerative (but not based on MSTs) Ward (1963) algorithm and to the one by Edwards and Cavalli-Sforza (1965) who employed an exhaustive divisive procedure.

**Table 1** Clustering methods studied

| | |
|---|---|
| *Agglomerative algorithms over MSTs* | |
| 1 | Single linkage (Florek et al., 1951) |
| 2–5 | Genie_G0.1, Genie_G0.3, Genie_G0.5, Genie_G0.7 (Genie with different Gini index thresholds; Algorithm 3; Gagolewski et al., 2016) |
| 6 | IcA (optimising the information criterion – starting from singletons; Algorithm 2) |
| 7–9 | Genie+Ic ($k+0$), Genie+Ic ($k+5$), Genie+Ic ($k+10$) (optimising the information criterion – agglomerative from a partial partition; Algorithm 4) |
| *Optimising different criteria – divisive strategy over MST (Algorithm 1)* | |
| 10 | ITM (information criterion; Müller et al., 2012) |
| 11 | MST/D_BallHall (the cluster validity index by Ball and Hall, 1965) |
| 12 | MST/D_CalinskiHarabasz (Caliński & Harabasz, 1974) |
| 13 | MST/D_DaviesBouldin (Davies & Bouldin, 1979) |
| 14–15 | MST/D_Silhouette, MST/D_SilhouetteW (average silhouette score and the mean of the cluster average silhouette widths; Rousseeuw, 1987) |
| 16–30 | MST/D_GDunn_d1_D1, MST/D_GDunn_d1_D2, MST/D_GDunn_d1_D3, MST/D_GDunn_d2_D1, MST/D_GDunn_d2_D2, MST/D_GDunn_d2_D3, MST/D_GDunn_d3_D1, MST/D_GDunn_d3_D2, MST/D_GDunn_d3_D3, MST/D_GDunn_d4_D1, MST/D_GDunn_d4_D2, MST/D_GDunn_d4_D3, MST/D_GDunn_d5_D1, MST/D_GDunn_d5_D2, MST/D_GDunn_d5_D3 (generalised Dunn (1974) indices; Bezdek and Pal, 1998) |
| 31–33 | MST/D_DuNN_25_Min_Max, MST/D_DuNN_25_Mean_Mean, MST/D_DuNN_-25_Max_Min (generalised Dunn indices based on near-neighbours; Gagolewski et al., 2021) |
| 34 | MST/D_WCNN_25 (within-cluster near-neighbour count; Gagolewski et al., 2021) |
| *Other algorithms based on MSTs* | |
| 35 | HEMST (Grygorash et al., 2006) |
| 36 | CTCEHC (Ma et al., 2021) |
| *Agglomerative hierarchical methods not based on MSTs* | |
| 37–42* | Average, complete, Ward, centroid, median, McQuitty linkages (e.g., Müllner, 2011) |
| 43* | Minimax (Bien & Tibshirani, 2011; Murtagh, 1983) |
| 44* | MinEnergy (Szekely & Rizzo, 2005) |
| 45–47* | HDBSCAN_4, HDBSCAN_2, HDBSCAN_8 (hierarchical density-based scan with different `minPts` settings; Campello et al., 2015) |
| *Other non-MST methods (`scikit-learn` package for Python;* Pedregosa et al., 2011) | |
| 48* | Gaussian Mixture (expectation-maximisation (EM) for Gaussian mixtures; `n_init=100`, `covariance_type="full"`; e.g., Dempster et al., 1977) |
| 49* | K-means (`n_init=100`; Lloyd, 1957) |
| 50* | Birch (`threshold=0.01`, `branching_factor=50`; Zhang et al., 1996) /best identified parameter setting/ |
| 51* | Spectral (K-means over a projection of the normalised Laplacian; `affinity="laplacian"` (kernel $K(x, y) = \exp(-\gamma\|x-y\|_1)$), $\gamma = 5$, `n_init=10`; e.g., Donath and Hoffman, 1973) /best identified parameter setting/ |
| *Other non-MST methods (via the `FCPS` package for R;* Thrun and Stier, 2021) | |
| 52* | Adaptive density peaks (Rodriguez & Laio, 2014; Wang & Xu, 2015) |
| 53* | Diana (DIvisive ANAlysis clustering; Rousseeuw and Kaufman, 1990) |
| 54* | Hardcl (On-line update algorithm – Hard competitive learning; Ripley, 2007) |
| 55* | Softcl (Neural gas – Soft competitive learning; Martinetz et al., 1993) |

**Table 1** continued

| | |
|---|---|
| 56* | Clara (Clustering LARge applications; Rousseeuw and Kaufman, 1990) |
| 57* | PAM (Partition around medoids; Rousseeuw and Kaufman, 1990) |
| *Other non-MST methods for reference (contribute to "Max Obs. Non-MST" in Table 3)* | |
| 58–109* | clusterings corresponding to maxima of 52 different cluster validity measures (Gagolewski et al., 2021) |
| 110–132* | Birch with 23 other parameter settings (`threshold` in {0.005, 0.01, 0.025, 0.05, 0.1, 0.25, 0.5, 1.0}, `branching_factor` in {10, 50, 100}; op.cit.) |
| 133–152* | Spectral with 19 other parameter settings (`affinity` in {"rbf", "laplacian", "poly", "sigmoid"} and $\gamma$ in {0.25, 0.5, 1.0, 2.5, 5.0}; op.cit.) |

Asterisks denote algorithms not based on MSTs

Let $F : \mathcal{X}_l \to \mathbb{R}$ be some objective function that we would like to maximise over the set of possible partitionings of any cardinality $l$ (not just $k$, which we treat as fixed). We will refer to it as a *cluster validity measure*. Moreover, given a subgraph $G' = (V, E'')$ of $G = (V, E)$ representing $\mathbf{X}$ such that $G'$ has $l$ connected components, let us denote by $C(V, E'') = (X_1, \ldots, X_l) \in \mathcal{X}_l$ the $l$-partition corresponding to these components.

**Algorithm 1** Maximising $F$ over an MST – Divisively

A general divisive scheme over an MST is a greedy optimisation algorithm that goes as follows:

1. Let $T = \mathrm{MST(G)} = (V, E', W')$;
2. Let $E'' = E'$;
3. For $i = 1, \ldots, k-1$ do:
   (a) Find $\{u, v\} \in E''$ which is a solution to:
   $$\max_{\{u,v\}} F(C(V, E'' \setminus \{\{u, v\}\}));$$
   (b) Remove $\{u, v\}$ from $E''$;
4. Return $C(V, E'')$ as a result.

In the single linkage scheme, the objective function is such that we maximise the sum of weights of the omitted MST edges. Furthermore, in the setting of the Caliński and Harabasz (1974) paper, we maximise (note the minus):

$$-\mathrm{WCSS}(X_1, \ldots, X_l) = -\sum_{i=1}^{l} \sum_{\mathbf{x}_j \in X_i} \|\mathbf{x}_j - \boldsymbol{\mu}_i\|^2, \tag{1}$$

where $\boldsymbol{\mu}_i$ is the centroid (componentwise arithmetic mean) of the $i$-th cluster. Naturally, many other objective functions can be studied. For instance, Müller et al. (2012) considered the information-theoretic criterion based on entropy[3] which takes into account cluster sizes and average within-cluster MST edges' weights:

$$\mathrm{IC}(X_1, \ldots, X_l) = -d \sum_{i=1}^{l} \frac{n_i}{n} \log \frac{L_i}{n_i} - \sum_{i=1}^{l} \frac{n_i}{n} \log \frac{n_i}{n}, \tag{2}$$

[3] Interestingly, the estimator given by Eq. 2 can be derived from the Rényi entropy approximation on various graph representations of data, not only MSTs (Eggels & Crommelin, 2019; Hero & Michel, 1998; Pál et al., 2010).

where $L_i$ denotes the sum of the weights of edges in the subtree of the MST representing the $i$-th cluster and $n_i$ denotes its size. This leads to an algorithm called ITM. Its Python implementation is available at https://github.com/amueller/information-theoretic-mst.

We can consider many internal cluster validity indices as objective functions. Each of them leads to a separate, standalone clustering algorithm (many of which have not yet been inspected in the literature). They are denoted by "MST/D_…" in Table 1. We will study the measures recently reviewed by Gagolewski et al. (2021):

- the indices by Ball and Hall (1965), Davies and Bouldin (1979, Def. 5), and Caliński and Harabasz (1974, Eq. 3) the latter being equivalent to the WCSS;
- two versions of the Silhouette index by Rousseeuw (1987);
- the generalisations of the Dunn (1974) index proposed by Bezdek and Pal (1998) (GDunn_dX_dY) and Gagolewski et al. (2021) (DuNN_M_X_Y);
- the within-cluster near-neighbour count (WCNN_M) by the same authors.

Let us stress that Gagolewski et al. (2021) considered the optimisation of these measures in the space of all possible partitions, whereas here, we restrict ourselves to the space induced by MSTs.

### 2.2 Agglomerative Algorithms over MSTs

The single linkage was rediscovered by Sneath (1957), who introduced it as a general agglomerative scheme. Its resemblance to the famous Kruskal (1956) MST algorithm (and hence that an MST suffices to compute it) was pointed out by, amongst others, Gower and Ross (1969). Starting from $n$ singletons, we keep merging the clusters containing the points joined by the edges of the MST in ascending order of weights. For a given MST with edges sorted increasingly, the amortised run-time is $O(n-k)$ based on the disjoint-sets (union-find) data structure (Cormen et al., 2009, Chapter 21).

Given a cluster validity measure $F$, we can generalise this agglomerative approach as follows.

---

**Algorithm 2** Maximising $F$ over an MST – agglomeratively.

A general agglomerative scheme over an MST is a greedy optimisation algorithm that consists of the following steps:

1. Let $T = \mathrm{MST}(\mathrm{G}) = (V, E', W')$;
2. Let $E'' = \emptyset$;
3. For $i = 1, \ldots, n-k$ do:
   (a) Find $\{u, v\} \in E' \setminus E''$ which is a solution to:
   $$\max_{\{u,v\}} F(C(V, E'' \cup \{\{u, v\}\}));$$
   (b) Add $\{u, v\}$ to $E''$;
4. Return $C(V, E'')$ as a result.

---

In the single linkage case, the objective is to maximise the sum of the weights of the unconsumed MST edges (edges in $E' \setminus (E'' \cup \{\{u, v\}\})$), or, equivalently, find the minimal $\{u, v\} \in E' \setminus E''$.

Unfortunately, the agglomerative procedure is oftentimes slow to compute; many cluster validity measures usually run in $\Omega(n^2)$ time and, as typically $k$ is much smaller than $n$, $O(n)$ iterations of the routine are required overall. Furthermore, the objective functions might not be well-defined for singleton and small clusters; this is problematic for they are the starting point of Algorithm 2. Hence, due to an already large number of procedures in our study, in the sequel, we will only consider the agglomerative maximising of the aforementioned information criterion (2), leading to the algorithm which we denote by IcA (information criterion – agglomerative strategy) in Table 1.

### 2.3 Variations on the Agglomerative Scheme

Genie proposed by Gagolewski et al. (2016) is an example variation on the agglomerative single linkage theme, where the total edge lengths are optimised in a greedy manner, but under the constraint that if the Gini index of the cluster sizes grows above a given threshold $g$, only the smallest clusters take part in the merging. Thanks to this, we can prevent the outliers from being classified as singleton clusters.

Formally, let $(c_1, \ldots, c_l)$ be a sequence such that $c_i$ denotes the cardinality of the $i$-th cluster in a given $l$-partition. The Gini index is defined as $\mathrm{G}(c_1, \ldots, c_l) = \sum_{i=1}^{l} (l - 2i + 1)c_{(i)}/(n-1)\sum_{i=1}^{l} c_i$, where $c_{(i)}$ denotes the $i$-th greatest value in the sequence. This index a measure of inequality of the cluster sizes: it takes the value of 0 if all components are equal, and yields 1 if all elements but one are equal to 0.

**Algorithm 3** Genie.

Given $g \in (0, 1]$:

1. Let $T = \mathrm{MST}(\mathrm{G}) = (V, E', W')$;
2. Let $E'' = \emptyset$;
3. For $i = 1, \ldots, n - k$ do:
   (a) If the Gini index of the sizes of clusters in $C(V, E'')$ is below $g$, pick $\{u, v\} \in E' \setminus E''$ as the edge with the smallest weight (equivalently, that the sum of weights of edges in $E' \setminus (E'' \cup \{\{u, v\}\})$ is the largest);
   (b) Otherwise, pick $\{u, v\} \in E' \setminus E''$ as the edge with the smallest weight provided that the size of the connected component containing $u$ (or $v$) is the smallest of them all;
   (c) Add $\{u, v\}$ to $E''$;
4. Return $C(V, E'')$ as a result.

We will rely on the implementation of Genie included in the `genieclust` package for Python and R (Gagolewski, 2021). Given a precomputed MST, the procedure runs in $O(n\sqrt{n})$ time.

The algorithm depends on the threshold parameter $g$. The original paper recommends the use of $g = 0.3$, but for some datasets, other settings may work better. In this study, we will only compare the results generated by $g \in \{0.1, 0.3, 0.5, 0.7\}$, as we observed that the clusterings obtained for the thresholds in-between these pivots do not differ substantially.

Cena (2018) in her PhD thesis, noted that the manual selection of the right $g$ threshold can be avoided by suggesting an agglomerative scheme optimising a cluster validity measure which uses a particular warm start. Namely, what we denote by Genie+Ic $(k + l)$ in Table 1, is a variation of Algorithm 2 maximising the information criterion (2), but one that starts at an intersection of clusters returned by the Genie algorithm with different parameters (as usually one of the thresholds leads to a clustering of high quality). In this paper, for greater flexibility, we allow the initial partitions to be possibly more fine-grained.

**Algorithm 4** Genie+Ic.

Given $l \geq 0$ and a set of $p$ thresholds $H = \{g_1, \ldots, g_p\}$ each in the unit interval, to find a $k$-clustering, proceed as follows:

1. Let $T = \mathrm{MST(G)} = (V, E', W')$;
2. For $j = 1, \ldots, p$, let $E''_j$ denote the final $E''$ from the run of Algorithm 3 with threshold $g_j$ seeking $k+l$ clusters;
3. Let $E'' = E' \setminus \bigcup_{j=1}^{p} E''_j$;
4. While $|E''| < n - k$:
   (a) Find $\{u, v\} \in E' \setminus E''$ which is a solution to:
   $$\max_{\{u,v\}} \mathrm{IC}(C(V, E'' \cup \{\{u, v\}\}));$$
   (b) Add $\{u, v\}$ to $E''$;
5. Return $C(V, E'')$ as a result.

We shall only consider $H = \{0.1, 0.3, 0.5, 0.7\}$ and $l \in \{0, 5, 10\}$, as we observe that other choices led to similar results. We have implemented this algorithm in the `genieclust` (Gagolewski, 2021) package.

Of course, any cluster validity index $F$ may be taken instead of IC in Algorithm 4. However, our preliminary research suggests that amongst many of the measures considered, the information criterion works best. Due to space constraints, these results will not be included here.

## 2.4 Other MST-Based Methods

Other MST-based methods that we implemented for the purpose of this study include:

- HEMST, which deletes edges from the MST to achieve the best attainable reduction in the standard deviation of edges' weights (Grygorash et al., 2006);
- CTCEHC, which constructs a preliminary partition based on the vertex degrees and then merges clusters based on the geodesic distance between the cluster centroids (Ma et al., 2021).

There are a few other MST-based methods in the literature, but usually they do not result in a given-in-advance number of clusters, $k$, a requirement which we impose in the next section for benchmarking purposes. For instance, Zahn (1971) determines the MST and deletes "inconsistent" edges (with weights significantly larger than the average weight of the nearby edges), but their number cannot be easily controlled.

We do not include the methods whose domain is not solely comprised of the information from MSTs (González-Barrios & Quiroz, 2003; Karypis et al., 1999; Mishra & Mohanty, 2019; Zhong et al., 2011, 2010). They construct MSTs based on transformed distances (Campello et al., 2015; Chaudhuri & Dasgupta, 2010), which use MSTs for very different purposes, such as auxiliary density estimation or some refinement thereof (Peter, 2013; Wang et al., 2009).

## 2.5 Reference Non-MST Methods

In the next section, we will compare the aforementioned algorithms against many popular non-MST approaches. They are denoted by an asterisk in Table 1. We consider two large open-source aggregator packages which provide consistent interfaces to many state-of-the-art methods. From `scikit-learn` in Python, see Pedregosa et al. (2011), we call the following:

- K-means following the Lloyd (1957) approach with `n_init=100` restarts from random initial candidate solutions, which is a heuristic that seeks cluster centres that minimise the within-cluster sum of squares;
- the expectation-maximisation algorithm for estimating the parameters of a Gaussian mixture with settings `n_init=100` and `covariance_type="full"`, i.e., each mixture component has its own general covariance matrix (e.g., Dempster et al. 1977);
- Spectral, which applies K-means (`n_init=10`) over a projection of the normalised Laplacian; amongst the 20 different parameter settings tested (`affinity` in {"rbf", "laplacian", "poly", "sigmoid"} and $\gamma$ in {0.25, 0.5, 1.0, 2.5, 5.0}), the best turned out to be `affinity="laplacian"`, i.e., based on the kernel $K(x, y) = \exp(-\gamma \|x - y\|_1)$ with $\gamma = 5$ (e.g., Donath and Hoffman, 1973);
- Birch proposed by Zhang et al. (1996), which incrementally builds the so-called clustering feature tree representing a simplified version of a given dataset (we tested 24 parameter sets: `threshold` in {0.005, 0.01, 0.025, 0.05, 0.1, 0.25, 0.5, 1.0} and `branching_factor` in {10, 50, 100}), and then clusters it using the Ward linkage algorithm; the best identified parameters were `threshold=0.01` and `branching_factor=50`;

whereas from `FCPS` in R, see Thrun and Stier (2021), we call:

- PAM (Partition around medoids), which is a robustified version of K-means that relies on the concept of cluster medoids rather than centroids (Rousseeuw & Kaufman, 1990);
- Clara (Clustering LARge Applications), which runs the foregoing PAM on multiple random dataset samples and then chooses the best medoid set (Rousseeuw & Kaufman, 1990);
- Diana (DIvisive ANAlysis clustering), which applies a divisive strategy to split clusters of the greatest diameters (Rousseeuw & Kaufman, 1990);
- Adaptive density peaks, which uses nonparametric multivariate kernels to identify areas of higher local point density and the corresponding cluster centres (Rodriguez & Laio, 2014; Wang & Xu, 2015);
- Hardcl (On-line update algorithm – Hard competitive learning) by Ripley (2007) and Softcl (Neural gas – Soft competitive learning) by Martinetz et al. (1993), both being a on-line version of K-means where centroids are iteratively moved towards randomly chosen points.

We also consider a few notable agglomerative hierarchical methods not based on MSTs:

- The classical agglomerative hierarchical clustering approaches utilising the average, complete, Ward, centroid, median, and McQuitty linkages implemented in the `fastcluster` package for Python by Müllner (2011);
- HDBSCAN by Campello et al. (2015) implemented in the `hdbscan` package for Python by McInnes et al. (2017), which uses a robustified version of the single linkage algorithm with respect to the so-called mutual reachability distance governed by different `minPts` parameter settings;

- Minimax, which is another agglomerative algorithm whose linkage function depends on the radius of a ball enclosing all points in the two clusters about to be merged (`protoclust` package for R); see the papers by Murtagh (1983) and Bien and Tibshirani (2011);
- MinEnergy (`energy` package for R), which is a variation on the Ward linkage proposed by Szekely and Rizzo (2005).

Furthermore, we include the clusterings that maximise 52 different cluster validity measures from the paper by Gagolewski et al. (2021).

We would like to emphasise that three of the above methods rely on feature engineering (selection of noteworthy features, scaling of columns, noise point or outlier removal, etc.), and were included as a point of reference for the algorithms that rely on "raw" distances. Spectral and Adaptive density peaks are kernel-based approaches which work in modified feature spaces, compare Wierzchoń and Kłopotek (2018) or Blum et al. (2020), and Gaussian Mixture can be thought of as a method that applies custom scaling of features in each cluster and thus does not rely on the original pointwise distance structure. Note that, overall, how to perform an appropriate feature engineering is a separate problem that should be carefully considered prior to an algorithm's run.

## 3 Experiments

### 3.1 Clustering Datasets, Reference Labels, and Assessing the Similarity Thereto

We have gathered many synthetic datasets from numerous sources, including those by renowned experts in the field, see the works by Fränti and Sieranoja (2018), Fränti and Virmajoki (2006), Sieranoja and Fränti (2019), Rezaei and Fränti (2016), Ultsch (2005), Thrun and Ultsch (2020), Thrun and Stier (2021), Graves and Pedrycz (2010), Dua and Graff (2021), Karypis et al. (1999), McInnes et al. (2017), Bezdek et al. (1999), and Jain and Law (2005), as well as instances that we have proposed ourselves. All datasets are included in the benchmark suite for clustering algorithms version 1.1.0 which is available for download from https://github.com/gagolews/clustering-data-v1/releases/tag/v1.1.0 (Gagolewski, 2022). Each of them comes with one or more reference label vectors.

For tractability reasons, we only take into account the datasets with $n < 10{,}000$. Moreover, we restrict ourselves to two- and three-dimensional instances ($d \leq 3$) so that the meaningfulness of the reference labels and the clustering results can be inspected visually. In total, we have 61 datasets; see Table 2 for the complete list.

In the "FCPS" and "SIPU" batteries, 25 partitions come from the original dataset creators, i.e., external experts: Thrun and Ultsch (2020), Fränti and Sieranoja (2018), etc. As van Mechelen et al. (2023) and Ullmann et al. (2022) advocate for minimising conflicts of interest in clustering evaluation, in the sequel, the subset consisting of these 25 "third-party" label sets will also be studied separately.

Other reference partitions were created by the current authors: in Table 2, we mark them clearly with an asterisk. We should note that even though the "Graves" battery includes datasets proposed by Graves and Pedrycz (2010), the original source did not define some of the reference partitions clearly enough (the figures therein are in greyscale). As most clusters therein are well-separable, we filled this gap to the best of our abilities. Similarly, the "chameleon_*" and "hdbscan" datasets in the "Other" battery were equipped with labels based on the points' density as we presumed that it was the intention of their original authors, Karypis et al. (1999) and McInnes et al. (2017).

**Table 2** Benchmark datasets studied ($n < 10{,}000$, $d \le 3$; see Sect. 3.1)

| | Battery | Dataset | $n$ | $d$ | $k$s |
|---|---|---|---|---|---|
| 1 | FCPS | atom | 800 | 3 | 2 |
| 2 | (Thrun & Ultsch, 2020) | chainlink | 1000 | 3 | 2 |
| 3 | | engytime | 4096 | 2 | 2, 2* |
| 4 | | hepta | 212 | 3 | 7 |
| 5 | | lsun | 400 | 2 | 3 |
| 6 | | target | 770 | 2 | 6, 2*$^{\dagger}$ |
| 7 | | tetra | 400 | 3 | 4 |
| 8 | | twodiamonds | 800 | 2 | 2 |
| 9 | | wingnut | 1016 | 2 | 2 |
| 10 | SIPU | a1 | 3000 | 2 | 20 |
| 11 | (Fränti & Sieranoja, 2018) etc | a2 | 5250 | 2 | 35 |
| 12 | | a3 | 7500 | 2 | 50 |
| 13 | | aggregation | 788 | 2 | 7 |
| 14 | | compound | 399 | 2 | 6, 4*, 5*$^{\dagger}$, 4*$^{\dagger}$, 5* |
| 15 | | d31 | 3100 | 2 | 31 |
| 16 | | flame | 240 | 2 | 2, 2*$^{\dagger}$ |
| 17 | | jain | 373 | 2 | 2 |
| 18 | | pathbased | 300 | 2 | 3, 4* |
| 19 | | r15 | 600 | 2 | 15, 9*, 8* |
| 20 | | s1 | 5000 | 2 | 15 |
| 21 | | s2 | 5000 | 2 | 15 |
| 22 | | s3 | 5000 | 2 | 15 |
| 23 | | s4 | 5000 | 2 | 15 |
| 24 | | spiral | 312 | 2 | 3 |
| 25 | | unbalance | 6500 | 2 | 8 |
| 26 | Graves | dense | 200 | 2 | 2* |
| 27 | (Graves & Pedrycz, 2010) | fuzzyx | 1000 | 2 | 5*, 2*$^{\dagger}$, 4*$^{\dagger}$, 2*$^{\dagger}$, 2*$^{\dagger}$ |
| 28 | | line | 250 | 2 | 2* |
| 29 | | parabolic | 1000 | 2 | 2*, 4* |
| 30 | | ring | 1000 | 2 | 2* |
| 31 | | ring_noisy | 1050 | 2 | 2*$^{\dagger}$ |
| 32 | | ring_outliers | 1030 | 2 | 5*, 2*$^{\dagger}$ |
| 33 | | zigzag | 250 | 2 | 3*, 5* |
| 34 | | zigzag_noisy | 300 | 2 | 3*$^{\dagger}$, 5*$^{\dagger}$ |
| 35 | | zigzag_outliers | 280 | 2 | 3*$^{\dagger}$, 5*$^{\dagger}$ |
| 36 | Other | chameleon_t4_8k | 8000 | 2 | 6*$^{\dagger}$ |
| 37 | (Karypis et al., 1999) | chameleon_t5_8k | 8000 | 2 | 6*$^{\dagger}$ |
| 38 | (McInnes et al., 2017) | chameleon_t8_8k | 8000 | 2 | 8*$^{\dagger}$ |
| 39 | | hdbscan | 2309 | 2 | 6*$^{\dagger}$ |
| 40 | | square | 1000 | 2 | 2* |

**Table 2** continued

| | Battery | Dataset | $n$ | $d$ | $k$s |
|---|---|---|---|---|---|
| 41 | WUT | circles* | 4000 | 2 | 4* |
| 42 | | cross* | 2000 | 2 | 4* |
| 43 | | graph* | 2500 | 2 | 10* |
| 44 | | isolation* | 9000 | 2 | 3* |
| 45 | | labirynth* | 3546 | 2 | 6* |
| 46 | | mk1* | 300 | 2 | 3* |
| 47 | | mk2* | 1000 | 2 | 2* |
| 48 | | mk3* | 600 | 3 | 3* |
| 49 | | mk4* | 1500 | 3 | 3* |
| 50 | | olympic* | 5000 | 2 | 5* |
| 51 | | smile* | 1000 | 2 | 6*, 4* |
| 52 | | stripes* | 5000 | 2 | 2* |
| 53 | | trapped_lovers* | 5000 | 3 | 3* |
| 54 | | twosplashes* | 400 | 2 | 2* |
| 55 | | windows* | 2977 | 2 | 5* |
| 56 | | x1* | 120 | 2 | 3* |
| 57 | | x2* | 120 | 2 | 3*, 4*† |
| 58 | | x3* | 185 | 2 | 4*, 3* |
| 59 | | z1* | 192 | 2 | 3* |
| 60 | | z2* | 900 | 2 | 5* |
| 61 | | z3* | 1000 | 2 | 4* |

Asterisks denote datasets or labellings generated by the current authors (labels provided by external experts are also studied separately). Daggers mark labellings which include noise points

Most importantly, the readers can independently inspect the sensibleness of the label vectors using the interactive data explorer tool available at https://clustering-benchmarks.gagolewski.com/. Many of the reference partitions are based on human Euclidean visual intuition, but some of them are also generated from different probability distribution mixtures. Overall, the benchmark datasets are quite diverse, which is important from the perspective of the results' generalisability and the fact that there exist different reasonable concepts of what a cluster is (Hennig, 2015; van Mechelen et al., 2023). Our battery includes well-separated clusters, overlapping point groups, clusters of various densities and shapes (Gaussian-like blobs, thin but long point chains, circles, etc.), noise points, and outliers.

The clustering algorithms were run in a purely unsupervised manner: they are only fed with the data matrix $\mathbf{X} \in \mathbb{R}^{n \times d}$ and $k$ as defined by the reference label vector, never with the particular true labels. As our aim is to study the behaviour of the algorithms on benchmark datasets of the form as proposed by their respective authors, no kind of data preprocessing (e.g., standardisation of variables, removal of noise points) was applied. Thus, all the algorithms under scrutiny were treated in the very same manner. In particular, all MST-based algorithms, K-means, and classical agglomerative methods (amongst others) rely on exactly the same information extracted from the raw Euclidean distance matrix. All of the computed partitions can be downloaded from and previewed at https://github.com/gagolews/clustering-results-v1.

As a measure of clustering quality, we consider the well-known adjusted Rand index (AR) by Hubert and Arabie (1985), which is given by:

$$\mathrm{AR}(\mathbf{C}) = \frac{\binom{n}{2}\sum_{i=1}^{k}\sum_{j=1}^{k}\binom{c_{i,j}}{2} - \sum_{i=1}^{k}\binom{c_{i,\cdot}}{2}\sum_{j=1}^{k}\binom{c_{\cdot,j}}{2}}{\binom{n}{2}\frac{1}{2}\left(\sum_{i=1}^{k}\binom{c_{i,\cdot}}{2} + \sum_{j=1}^{k}\binom{c_{\cdot,j}}{2}\right) - \sum_{i=1}^{k}\binom{c_{i,\cdot}}{2}\sum_{j=1}^{k}\binom{c_{\cdot,j}}{2}}, \tag{3}$$

where the confusion matrix $\mathbf{C}$ is such that $c_{i,j}$ denotes the number of points in the $i$-th reference cluster that a given algorithm assigned to the $j$-th cluster, with $c_{i,\cdot} = \sum_{j=1}^{k} c_{i,j}$, and $c_{\cdot,j} = \sum_{i=1}^{k} c_{i,j}$. AR counts the number of concordant pairs between two clusterings and is adjusted for chance: its expected value is 0 if two independent partitions generated from the same marginal distribution are given. The index yields the value of 1 only if two partitions are identical, and a negative value for partitions "worse than random". For more properties of AR, see, e.g., the work by Warrens and van der Hoef (2022).

Some datasets come with many reference label vectors: the total number of unique partitions is 81. This is in line with the recommendations by Gagolewski (2022), see also Dasgupta and Ng (2009) and von Luxburg et al. (2012), who argued that there can be many possible ways to cluster a dataset and thus an algorithm should be rewarded for finding any of the reference partitions. Thus, for each clustering algorithm and dataset, the maximum of the adjusted Rand indices with respect to all reference label vectors was considered (e.g., for Graves/fuzzy, we take the maximum of five similarity scores).

It is also worth noting that 19 partitions include some points marked as noise: they do not belong to any actual cluster, but are supposed to make an algorithm's job more difficult. They are ignored when computing the confusion matrix after the clustering has been performed.

If an algorithm failed to converge or generated an error, we set the corresponding AR to 0. This was the case only in six instances (twice for spectral and once for Softcl, MST/D_WCNN_25, MST/D_DuNN_25_Mean_Mean, and MST/D_DuNN_25_Min_Max).

### 3.2 Are MST-Based Methods Promising?

Recall that the number of possible partitions of an MST with $n$ edges into $k$ subtrees is equal to $(n-1)(n-2)\cdots(n-k+1)$. For all datasets with $k \le 4$, we identified the true maximum of AR using the brute-force approach, i.e., considering all the possible partitions generated by the cancelling of edges from the MST. The remaining cases were too time-consuming to examine exhaustively. Therefore, we applied a tabu-like steepest ascent search strategy with at least 10 random restarts. Although there is no theoretical guarantee that the identified maxima are global, the results are promising as they agreed with the exact solutions for $k \le 4$ and were no worse than the observable maximal AR scores for the MST-based methods.

Let us thus consider the "Max MST" rows in Table 3, which denote the aforementioned theoretically achievable maximal AR index values (a hypothetical "oracle" MST-based algorithm). We see that overall, MST-based methods are potentially very competitive, despite their simplicity. Only in $6/61 \simeq 10\%$ instances ($2/25 = 8\%$ if we restrict ourselves to third-party labels), we obtained the AR index less than 0.8, namely: SIPU/s3, s4, WUT/z1, graph, twosplashes, and olympic (see Fig. 2 for the depiction of the last two datasets). Here, the minimum spanning trees with respect to the Euclidean distance between unpreprocessed points do not constitute valid representations of the feature space, as the clusters are too overlapping. Also, $10/61 \simeq 16\%$ ($5/25 \simeq 20\%$) cases end up with AR $< 0.95$.

Nevertheless, we must note that some datasets are problematic (AR $< 0.8$) also for non-MST methods that we described in Sect. 3.4 (compare the "Max Obs. Non-MST" rows in

**Table 3** Aggregated AR indices across all the algorithms: the number of datasets where AR $< 0.8$, AR $\geq 0.95$, as well as the minimum, the 1st quartile, median, and arithmetic mean of AR

| | $< .8$ | $\geq .95$ | Min | Q1 | Med | Mean |
|---|---|---|---|---|---|---|
| Third-party labels only (25 instances) | | | | | | |
| Max All | 2 | 20 | .65 | .97 | 1 | .96 |
| Max MST | 2 | 20 | .65 | .96 | 1 | .95 |
| Max Obs | 3 | 19 | .64 | .95 | 1 | .94 |
| Max Obs. MST | 4 | 16 | .62 | .94 | .99 | .93 |
| Max Obs. Non-MST | 3 | 19 | .61 | .95 | 1 | .94 |
| All labels (61 instances) | | | | | | |
| Max All | 3 | 53 | .65 | 1 | 1 | .97 |
| Max MST | 6 | 51 | .27 | .99 | 1 | .95 |
| Max Obs | 4 | 51 | .64 | .98 | 1 | .97 |
| Max Obs. MST | 7 | 44 | .23 | .95 | 1 | .93 |
| Max Obs. Non-MST | 7 | 50 | .49 | .98 | 1 | .95 |

"Max Obs." gives the maximal observed AR based on the outputs of all the methods listed in Table 1, and their counterparts for the MST and non-MST algorithms only are denoted by "Max Obs. MST" and "Max Obs. Non-MST". "Max MST" gives the theoretically achievable maxima of the accuracy scores for the MST-based methods. Moreover, "Max All" is the maximum of "Max MST" and "Max Obs."

Table 3). These are: SIPU/s3, s4, pathbased, Other/chameleon_t8_8k, WUT/cross, twosplashes, and Graves/parabolic.

Overall (compare "Max All"), $53/61 \simeq 87\%$ ($20/25 \simeq 75\%$) instances are discovered correctly (AR $\geq 0.95$) by at least one clustering approach; more precisely, all but SIPU/s2, s3, s4, WUT/twosplashes, graph, mk3, olympic, and Graves/parabolic. These eight failures might mean that:

- better clustering algorithms should be developed or taken into account,
- some further data preprocessing must be applied in order to reveal the cluster structure; this is true for the WUT/twosplashes dataset which benefits from the standardisation of features,
- there is something wrong with the reference label vectors, e.g., an alternative sensible partition exists but was not formalised, the clusters are overlapping too heavily in which case the partitional clustering algorithms should be replaced by soft/fuzzy ones.

### 3.3 Which MST-Based Algorithms Work Best?

That the MST-based methods are generally promising does not mean that we are in possession of an algorithm that gets the most out of the information conveyed by the minimum spanning trees, nor that a single strategy is always best. We should thus inspect which approaches and/or objective functions are more useful than others.

Figure 3 depicts, for each MST-based method, the adjusted Rand indices relative to the aforementioned "Max MST", i.e., how well each algorithm compares to the best theoretically achievable solution.

The agglomerative Genie algorithm is the top performer. The approaches optimising the information criterion (Genie+Ic, ITM) also give high average relative AR. Optimising the recently-proposed near-neighbour-based index, DuNN_25_Min_Max, yields high median relative scores too.

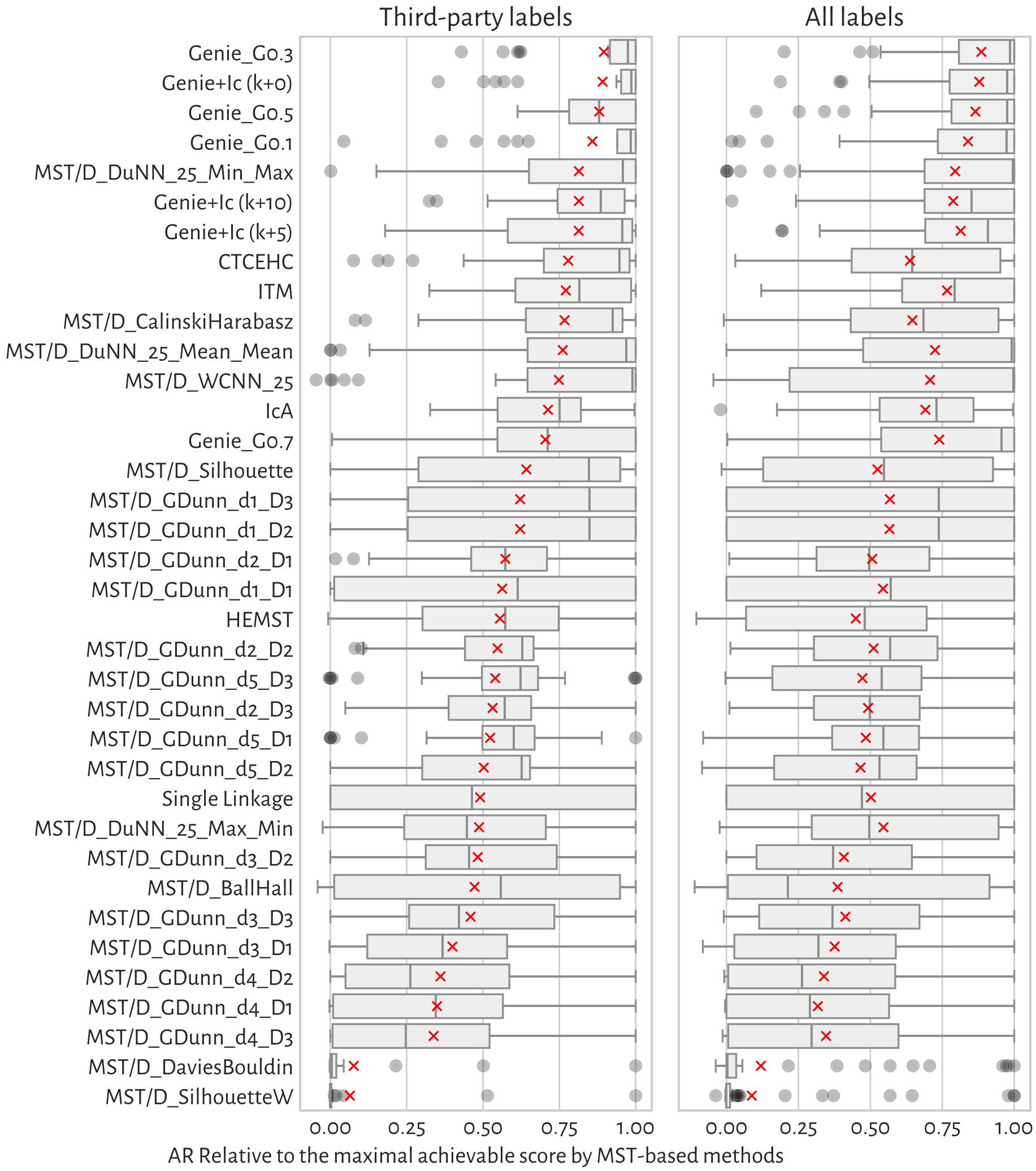

**Fig. 3** The distribution of the adjusted Rand indices for all MST-based algorithms relative to the "Max MST" AR score (see Sect. 3.2), ordered with respect to the average relative AR (red crosses) in the case of the third-party labels

As far as other "standalone" algorithms are concerned, HEMST and Single linkage exhibit inferior performance, and CTCEHC is comparable with the divisive Caliński–Harabasz criterion optimiser.

Quite strikingly, some well-established internal cluster validity measures promote clusterings of very poor agreeableness with the reference labels (Davies–Bouldin, SilhouetteW, some generalised Dunn indices). Choosing the wrong objective function to optimise over MST can lead to very poor results. This puts their meaningfulness into question. This is in line with the observation made by Gagolewski et al. (2021), where a similar study over the space of all possible partitionings was performed.

### 3.4 How MST-Based Methods Compare Against Other Clustering Approaches?

Let us compare the best MST-based approaches with other popular clustering algorithms mentioned in Sect. 2.5. Table 4 gives the summary of the AR indices over all 61 benchmark datasets and their subset consisting of 25 third-party labels only. In both cases, the best MST-based methods are high-performers: Genie_G0.3, Genie+Ic (k+0), and MST/D_DuNN_25_Min_Max identify approximately half of the reference partitions correctly (AR $\geq 0.95$).

**Table 4** Aggregated adjusted Rand indices for different algorithms (best results marked in bold)

| | $< .8$ | $\geq .95$ | Med | Mean | $< .8$ | $\geq .95$ | Med | Mean |
|---|---|---|---|---|---|---|---|---|
| | Third-party labels only (25) | | | | All labels (61 instances) | | | |
| Genie_G0.3 | **7** | **12** | **0.94** | **0.85** | **18** | **35** | **0.99** | **0.85** |
| Genie+Ic (k+0) | **7** | **13** | **0.96** | **0.85** | **19** | **32** | **0.96** | **0.84** |
| Spectral* | **8** | **13** | **0.96** | **0.83** | 30 | 24 | 0.86 | 0.69 |
| MST/D_DuNN_25_Min_Max | **8** | **12** | 0.92 | 0.79 | **20** | **35** | **0.99** | 0.78 |
| Gaussian Mixture* | **9** | 10 | **0.94** | 0.75 | 28 | 22 | 0.86 | 0.67 |
| MinEnergy* | 12 | 8 | 0.90 | 0.75 | 38 | 15 | 0.65 | 0.61 |
| CTCEHC | 10 | 8 | 0.89 | 0.74 | 40 | 11 | 0.63 | 0.61 |
| MST/D_WCNN_25 | **9** | **12** | **0.94** | 0.74 | 22 | **33** | 0.99 | 0.70 |
| ITM | 16 | 7 | 0.77 | 0.73 | 37 | 18 | 0.75 | 0.73 |
| MST/D_CalinskiHarabasz | 12 | 5 | 0.84 | 0.73 | 36 | 11 | 0.62 | 0.61 |
| K-means* | 12 | 9 | 0.82 | 0.71 | 38 | 17 | 0.61 | 0.58 |
| PAM* | 13 | 10 | 0.77 | 0.70 | 41 | 15 | 0.61 | 0.57 |
| Birch* | 11 | 8 | 0.83 | 0.70 | 36 | 16 | 0.62 | 0.58 |
| Adaptive Density Peaks* | 13 | 8 | 0.76 | 0.70 | 41 | 12 | 0.59 | 0.56 |
| Clara* | 13 | 6 | 0.75 | 0.70 | 41 | 11 | 0.62 | 0.57 |
| Average Linkage* | 11 | 8 | 0.91 | 0.68 | 39 | 15 | 0.53 | 0.54 |
| Ward Linkage* | 13 | 6 | 0.72 | 0.68 | 38 | 13 | 0.62 | 0.57 |
| IcA | 20 | 4 | 0.72 | 0.68 | 45 | 8 | 0.70 | 0.65 |
| Centroid Linkage* | 11 | 8 | 0.87 | 0.65 | 39 | 15 | 0.51 | 0.53 |
| Softcl* | 14 | 5 | 0.73 | 0.64 | 39 | 13 | 0.59 | 0.56 |
| Minimax* | 13 | 7 | 0.77 | 0.64 | 41 | 11 | 0.45 | 0.51 |
| Hardcl* | 16 | 2 | 0.67 | 0.63 | 44 | 6 | 0.58 | 0.53 |
| Complete Linkage* | 15 | 6 | 0.78 | 0.63 | 43 | 10 | 0.44 | 0.52 |
| Diana* | 14 | 1 | 0.79 | 0.62 | 41 | 7 | 0.53 | 0.53 |
| MST/D_Silhouette | 12 | 4 | 0.81 | 0.61 | 40 | 9 | 0.46 | 0.50 |
| McQuitty Linkage* | 16 | 4 | 0.61 | 0.59 | 45 | 7 | 0.43 | 0.48 |
| Median Linkage* | 16 | 5 | 0.71 | 0.56 | 48 | 7 | 0.39 | 0.45 |
| HEMST | 20 | 4 | 0.55 | 0.54 | 50 | 8 | 0.45 | 0.43 |
| HDBSCAN_2* | 16 | 8 | 0.44 | 0.48 | 36 | 23 | 0.47 | 0.50 |
| Single Linkage | 16 | 8 | 0.44 | 0.48 | 36 | 23 | 0.47 | 0.50 |
| HDBSCAN_4* | 18 | 6 | 0.15 | 0.37 | 39 | 21 | 0.18 | 0.44 |
| HDBSCAN_8* | 21 | 3 | 0.01 | 0.23 | 45 | 15 | 0.01 | 0.31 |

Asterisks denote algorithms not based on MSTs

The spectral and Gaussian mixture methods (which are not based on MSTs) have some potential leverage as they apply a form of feature engineering automatically. Despite this, they did not outperform their competitors, but overall they work very well too. However, their performance drops when we use the extended dataset repertoire that includes cases of noisy clusters and those of many different shapes. In particular, even if we exclude the WUT datasets created by a subset of the current authors, the "Graves" and "Other" batteries (which were proposed by other researches) feature 1 "difficult" (AR $< 0.8$) and 13 "easy" (AR $\geq 0.95$) cases for the Genie_G0.3 algorithm, and 8 "difficult" but only 5 "easy" instances for the spectral one. Overall, the average AR indices on the 40 first datasets from Table 2 are equal to 0.9 (increase from 0.85) and 0.79 (decrease from 0.83) for Genie_G0.3 and spectral, respectively.

On the other hand, the fact that the average and centroid linkages, K-means, and PAM methods were able to identify 8–10 partitions from the "SIPU" and "FCPS" subsets may indicate that those batteries consist of many datasets with isotropic Gaussian-like blob-shaped clusters.

Certain more recent methods: MinEnergy, CTCEHC, and ITM perform reasonably well. We should also take note of the quite disappointing performance of the HDBSCAN algorithm, which is essentially the (robustified) single linkage method applied on a version of the distance matrix that lessens the influence of potential outliers (where the `minPts` parameter plays the role of a smoothing factor). HDBSCAN is, theoretically, capable of marking some points as noise, but it did not do so on our benchmark data.

## 4 Conclusion

Naturally, no algorithm can perform best in all the scenarios. Hence, in practice, it is worth to consider the outputs of many methods at the same time. We demonstrated that the minimum spanning tree-based methods are potentially very competitive clustering approaches. MSTs are suitable for representing dense clusters of arbitrary shapes, and can be relatively robust to outliers in the outer parts of the feature space.

Many MST-based methods are quite simple and easy to compute: once the minimum spanning tree is determined (which takes up to $O(n^2)$ time, but approximate methods exist too; e.g., Naidan et al. 2019), we can potentially get a whole hierarchy of clusters of any cardinality. This property makes them suitable for solving extreme clustering tasks, where the number of clusters $k$ is large (Kobren et al., 2017). For instance, the Genie algorithm, which is the top performer in this study, needs $O(n\sqrt{n})$ time, given a prebuilt MST, to generate partitions of all cardinalities. On the other hand, the well-known K-means algorithm is fast for small fixed $k$s only as each iteration thereof requires $O(ndk)$ time.

Although MSTs can be distorted by introducing many noise points in-between the well-separable clusters, they can most likely be improved by appropriate feature engineering, e.g., scaling of data columns, noise point removal, modifying the distance matrix, automatic data projections onto simpler spaces, etc. (Blum et al., 2020; Campello et al., 2015; D'Urso & Vitale, 2022; Temple, 2023; Yin & Liu, 2009). Combining automated data preprocessing with clustering leads to a distinct class of methods: they should be studied separately, so we leave them as a topic for future research.

Overall, no single MST-based method surpasses all others, but the new divisive and agglomerative approaches we have proposed in this paper perform well on certain dataset types. However, compared to the theoretically achievable maxima, we note that there is still some room for improvement (see Table 3). Thus, the development of new algorithms is encouraged. In particular, it might be promising to explore the many possible combinations of parameters and objective functions we have left out due to the obvious space constraints in this paper.

Future work should involve the testing of clustering methods based on near-neighbour graphs and more complex MST-inspired data structures (Fränti et al., 2006; González-Barrios & Quiroz, 2003; Karypis et al., 1999; Zhong et al., 2011, 2010). Also, the application of the MST-based algorithms could be examined in the problem of community detection in graphs (Gerald et al., 2023).

It would be also interesting to inspect the stability of the results when different random subsets of benchmark data are selected or study the problem of overlapping clusters (Campagner et al., 2023). Moreover, let us again emphasise that have studied only low-dimensional datasets ($d \leq 3$) so that the results could be assessed visually. A separate analysis, requiring most likely a different methodology, should be performed on medium- and large-dimensional data.

Finally, let us recall that we have only focused on partitional clustering methods that guarantee to return a fixed-in-advance number of clusters $k$. In the future, it would be interesting to allow for the relaxation of these constraints, e.g., study cluster hierarchies and soft clustering approaches.

**Acknowledgements** The authors would like to thank the Editor and Reviewers for providing their helpful comments that led to the improvement of earlier versions of this paper.

**Author Contributions** MG: conceptualisation, methodology, data curation, software, visualisation, investigation, formal analysis, writing — original draft, writing — revision; AC: methodology, data curation, investigation; MB: software, data curation, investigation; ŁB: software, investigation

**Funding** This research was partially supported by the Australian Research Council Discovery Project ARC DP210100227 (MG).

**Data Availability** All benchmark data are publicly available at https://clustering-benchmarks.gagolewski.com/ (Gagolewski, 2022). The project homepage features a visual dataset explorer. We used version 1.1.0 of the benchmark dataset battery, which can be fetched from https://github.com/gagolews/clustering-data-v1/releases/tag/v1.1.0.
All computed partitions can be downloaded from and previewed at https://github.com/gagolews/clustering-results-v1.
All cluster validity measures are implemented in the `genieclust` package for Python and R (Gagolewski, 2021). Our implementation of the HEMST and CTCEHC algorithms is available at https://github.com/lukaszbrzozowski/msts.

## Declarations

**Conflict of Interest** The authors declare no competing interests.

**Publisher's Note** Springer Nature remains neutral with regard to jurisdictional claims in published maps and institutional affiliations.